\documentclass[review]{elsarticle}
\usepackage{graphicx}
\usepackage{amssymb}
\usepackage{amsmath}

\usepackage{lineno}
\journal{arXiv (version 1)}
\begin{document}
\begin{frontmatter}
\title{Endoscopy artifact detection (EAD 2019) challenge dataset}
\author[1]{Sharib Ali\corref{cor1}}
\ead{sharib.ali@eng.ox.ac.uk}
\author[2]{Felix Zhou}
\author[3]{Christian Daul}
\author[4]{Barbara Braden}
\author[4]{Adam Bailey}
\author[5]{Stefano Realdon}
\author[4]{James East}
\author[12]{Georges Wagni\`{e}res}
\author[13]{Victor Loschenov}
\author[14]{Enrico Grisan}
\author[3]{Walter Blondel}
\author[1]{Jens Rittscher}
\address[1]{Institute of Biomedical Engineering and Big Data Institute, Old Road Campus, University of Oxford, Oxford, UK}
\address[2]{Ludwig Institute for Cancer Research, University of Oxford, Oxford, Oxford, UK}
\address[3]{CRAN UMR 7039, University of Lorraine, CNRS, Nancy, France}
\address[4]{Translational Gastroenterology Unit, Experimental Medicine Div., John Radcliffe Hospital, University of Oxford, Oxford, UK}
\address[5]{Instituto Onclologico Veneto, IOV-IRCCS, Padova, Italy}
\address[12]{Swiss Federal Institute of Technology at Lausanne (EPFL), Lausanne, Switzerland}
\address[13]{Prokhorov General Physiscs Institute of the Russian Academy of Sciences, Moscow, Russia}
\address[14]{Department of Information Engineering, University of Padova, Padova, Italy}
\cortext[cor1]{Sharib Ali}
\begin{abstract}
Endoscopic artifacts are a core challenge in facilitating the diagnosis and treatment of diseases in hollow organs. Precise detection of specific artifacts like pixel saturations, motion blur, specular reflections, bubbles and debris is essential for high-quality frame restoration and is crucial for realizing reliable computer-assisted tools for improved patient care. At present most videos in endoscopy are currently not analyzed due to the abundant presence of multi-class artifacts in video frames. Through the endoscopic artifact detection (EAD 2019) challenge, we address this key bottleneck problem by solving the
accurate identification and localization of endoscopic frame artifacts to enable further key quantitative analysis of unusable video frames such as mosaicking and 3D reconstruction which is crucial for delivering improved patient care. This paper summarizes the challenge tasks and describes the dataset and evaluation criteria established in the EAD 2019 challenge. 
\end{abstract}
\end{frontmatter}
\section{Introduction}
\label{S:1}
Endoscopy is a widely used clinical procedure for the early detection of numerous cancers (e.g., nasopharyngeal, oesophageal adenocarcinoma, gastric, colorectal cancers, bladder cancer etc.), therapeutic procedures and minimally invasive surgery (e.g., laparoscopy). During this procedure an endoscope is used; a long, thin, rigid or flexible tube having a light source and a camera at the tip which allows to visualize inside of affected organs on a screen. A major drawback of these video frames is the heavy corruption of multiple imaging artifacts (e.g., pixel saturation, motion blur, defocus blur, specular reflections, bubbles, fluid and debris). These artifacts not only present difficulty in visualizing the underlying tissue during diagnosis but also adversely affects post-analysis methods. Many notable quantitative analysis, such as video mosaicking for visualizing an extended field-of-view for follow up, 3D surface reconstruction for aiding surgical planning and key video-frame retrieval for reporting, that help to significantly improve clinical care are compromised. The accurate detection of artifacts in clinical endoscopy is thus a critical bottleneck problem whose solution will transform and greatly accelerate the development of effective quantitative clinical endoscopic analysis across all diseases, organs and modalities. This was first identified in our previous work~\cite{Ali_arXiv2019}. However, to comprehensively address the artifact detection problem and stimulate academic discussion, we established the Endoscopy artifact Detection challenge (EAD)\footnote{Details of this challenge can be found at: https://ead2019.grand-challenge.org} as an initiative to discover the
limitations of existing state-of-the-art computer vision methods and to stimulate the development of new algorithms in the following key problem areas inherent to all video endoscopy.

\subsection{Multi-class artifact detection}
Existing endoscopy workflows detect mainly one artifact class which is insufficient to obtain high-quality frame restoration suitable for quantitative analysis of the entire clinical video. In general, the same video frame can be corrupted with multiple artifacts for example motion blur, specular reflections, and low contrast can be present in the same frame. Further, not all artifact types corrupt the frame to equal extent. Thus unless multiple artifacts present in the frame are known with their precise spatial location, clinically relevant frame restoration quality cannot be guaranteed. Another advantage of class specific detection is that frame quality assessments can be more guided to minimise the overall number of frames that gets discarded during automated video analysis maximising the usage of information within each video.

\subsection{Multi-class artifact region segmentation: }
Frame artifacts typically have irregular shapes that are non-rectangular. Consequently they are overestimated by bounding box detections. The development of accurate semantic segmentation methods to precisely delineate the boundaries of each detected frame artifact enables optimized restoration of video frames without sacrificing information.

\subsection{Multi-class artifact generalisation}
It is important for algorithms to avoid biases induced by the use of specific training datasets. Additionally it is well known that expert annotation generation is time consuming and infeasible for many institutions. In this challenge, we encourage the participants to develop machine learning algorithms that can be applied across different endoscopic datasets worldwide based on our large collected combined dataset from 6 different institutions. 

\section{Dataset}
\begin{figure}[t!]
    \centering
    \includegraphics[scale=0.33]{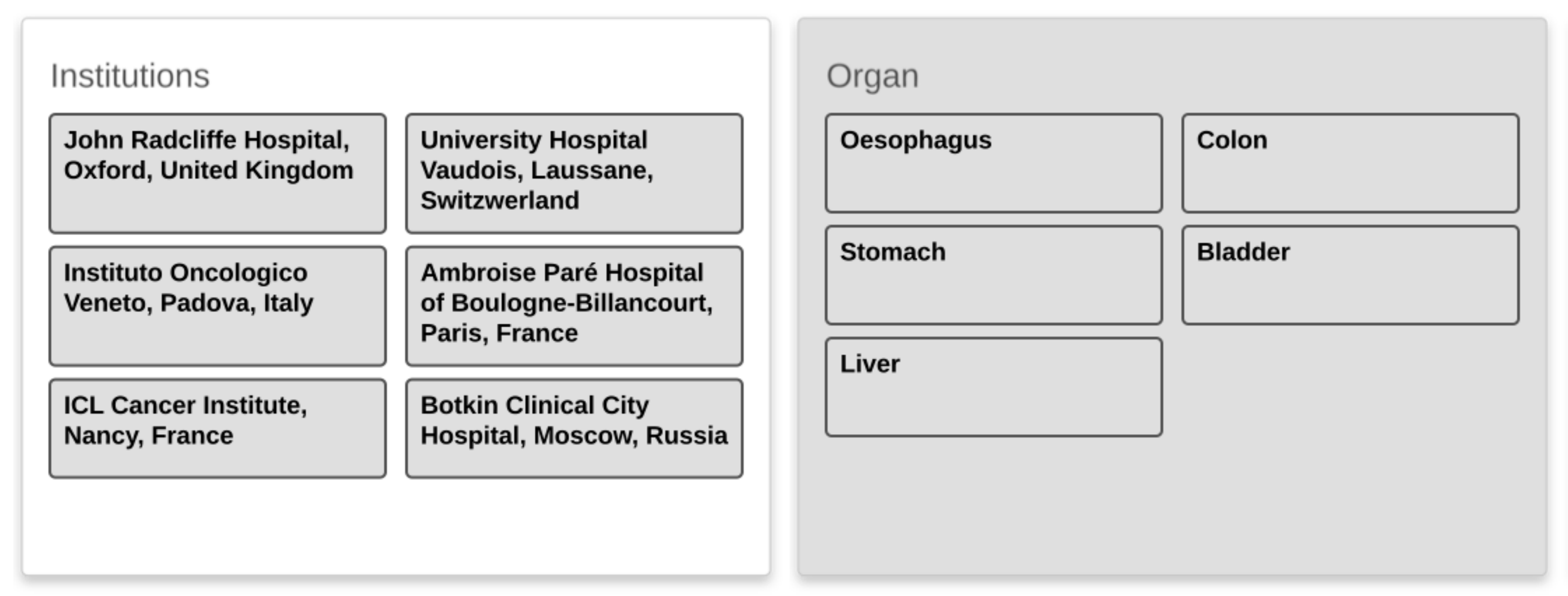}
 \includegraphics[scale=0.3]{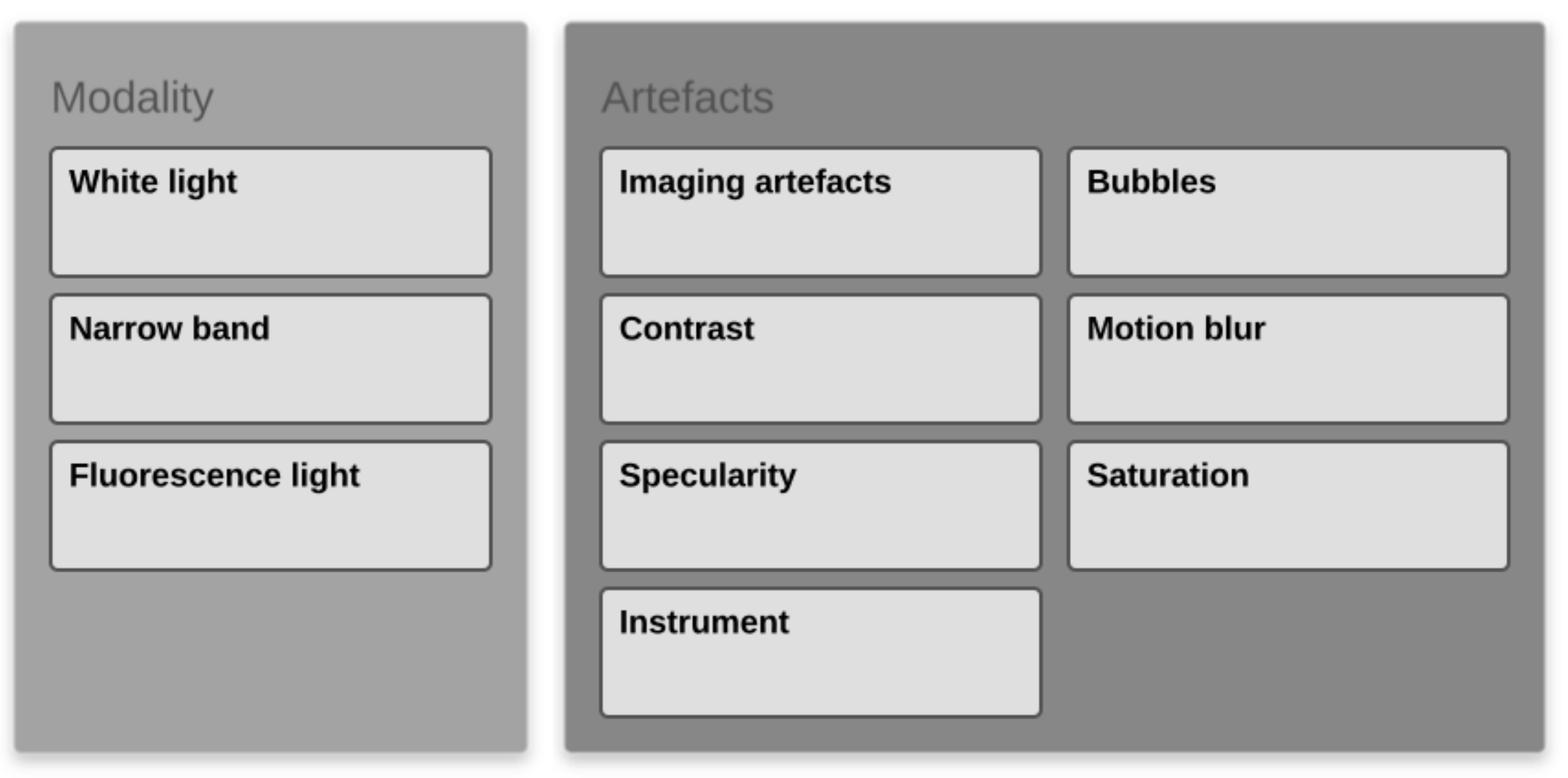}
    \caption{Summary of EAD2019 challenge dataset cohort.}
    \label{fig:ead2019_datasummary}
\end{figure}

With the EAD Challenge we aimed to establish a first large and comprehensive dataset for ``Endoscopy artifact detection'' (see Fig.~\ref{fig:ead2019_datasummary}). The provided data was assembled from 6 different centers worldwide: John Radcliffe Hospital, Oxford, UK; ICL Cancer Institute, Nancy, France; Ambroise Par\`{e} Hospital of Boulogne-Billancourt, Paris, France; Instituto Oncologico Veneto, Padova, Italy; University Hospital Vaudois, Lausanne, Switzerland and the Botkin Clinical City Hospital, Moscow. This unique endoscopic video frame dataset is multi-tissue (gastroscopy, cystoscopy, gastrooesophageal, colonoscopy), multi-modal (white light, fluorescence, and narrow band imaging), is inter patient and encompasses multiple populations (UK, France, Russia, and Switzerland). Videos were collected from patients on a first-come-first-served basis at Oxford, with randomized sampling at French centres and only cancer patients were selected at the Moscow centre. Videos at these centres were acquired with standard imaging protocols using endoscopes built by different companies, Olympus, Biospec, and Karl Storz. The dataset was built randomly mixing the collected data with no exclusion criteria. All images have been carefully anonymised. No patient information should be visible in this data. A comprehensive open-source software\footnote{Useful tools for this dataset: https://sharibox.github.io/EAD2019/} have been established to assist the participants.
\subsection{Gold standard}
Clinical relevance to the challenge problem was first identified. During this step, 7 (see Fig.~\ref{fig:ead2019_datasummary}) different common imaging artifact types were suggested by 2 expert clinicians who performed bounding box labelling of these artifacts on a small dataset ($\sim$100 frames). These frames were then taken as reference to produce bounding box annotations for the remaining train-test dataset by 2 experienced postdoctoral fellows. Finally, further validation by 2 experts (clinical endoscopists) was carried out to ensure the reference standard. The ground-truth labels were randomly sampled (1 per 20 frames) during this process.  

To maximise consistency in annotation labels between annotators a few rules were determined as described below. For the final scoring, we additionally penalized annotator variance of IoU (intersection over union, in the final score) which exhibit inevitable bounding box variations between annotators particularly for more subjective artifact types such as blur and contrast. 

\begin{figure}[t!]
\centering
\begin{minipage}[b]{0.285\linewidth}
\includegraphics[width=\linewidth]{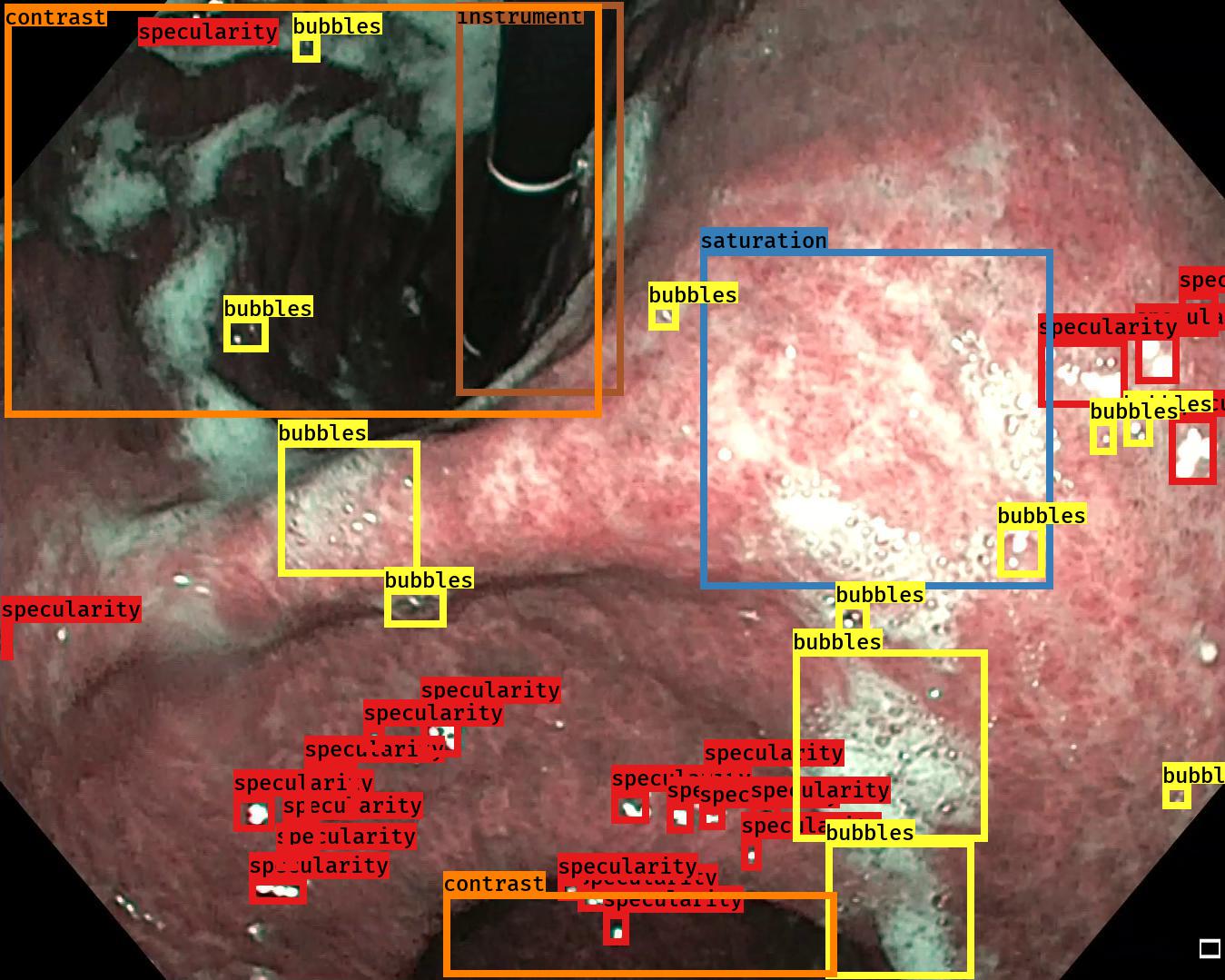}
\end{minipage}
\begin{minipage}[b]{0.285\linewidth}
\includegraphics[width=\linewidth]{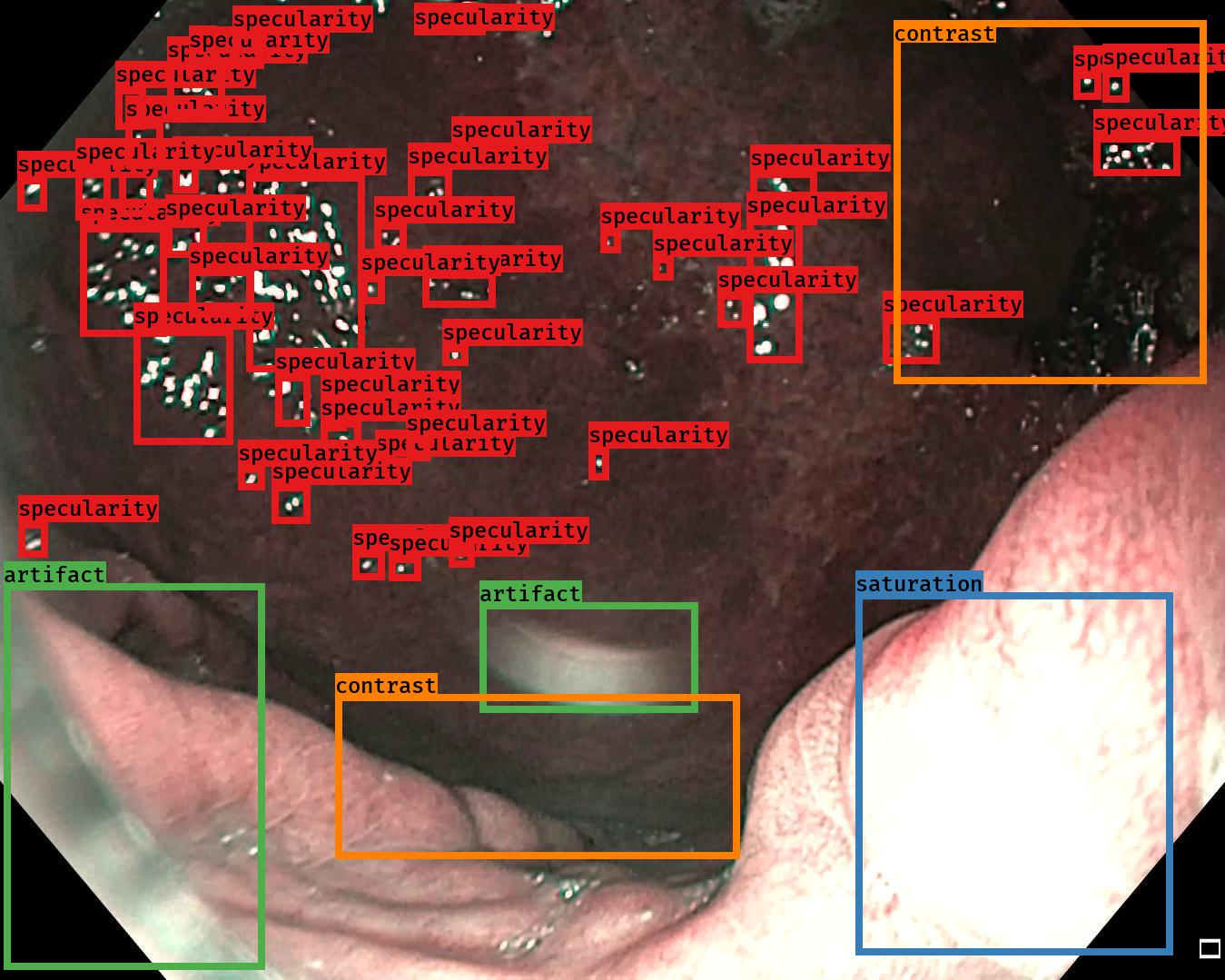}
\end{minipage}
\begin{minipage}[b]{0.24\linewidth}
\includegraphics[width=\linewidth]{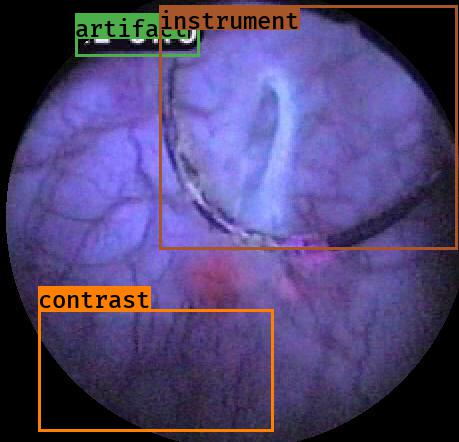}
\end{minipage}
\begin{minipage}[b]{0.32\linewidth}
\includegraphics[width=\linewidth]{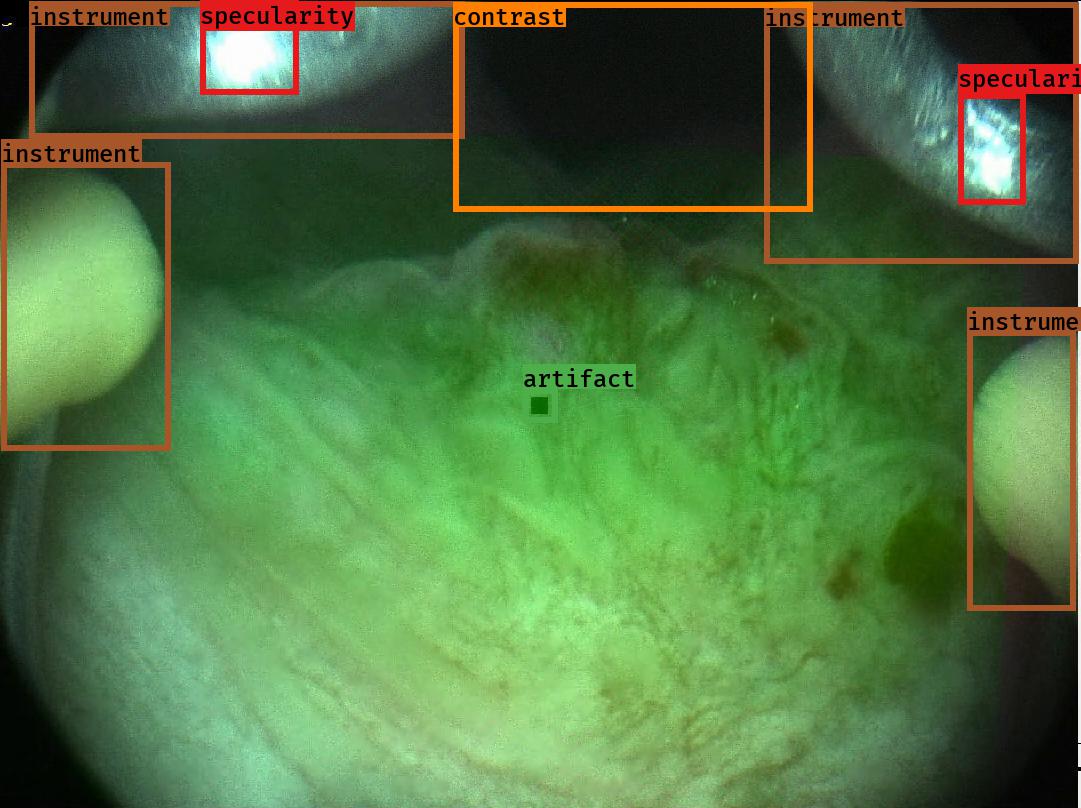}
\end{minipage}
\begin{minipage}[b]{0.24\linewidth}
\includegraphics[width=\linewidth]{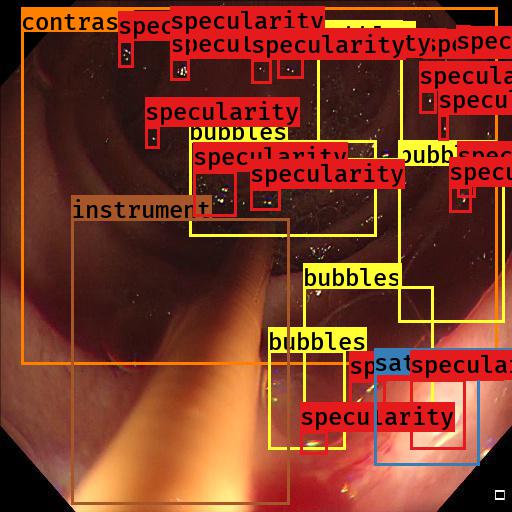}
\end{minipage}
\begin{minipage}[b]{0.3\linewidth}
\includegraphics[width=\linewidth]{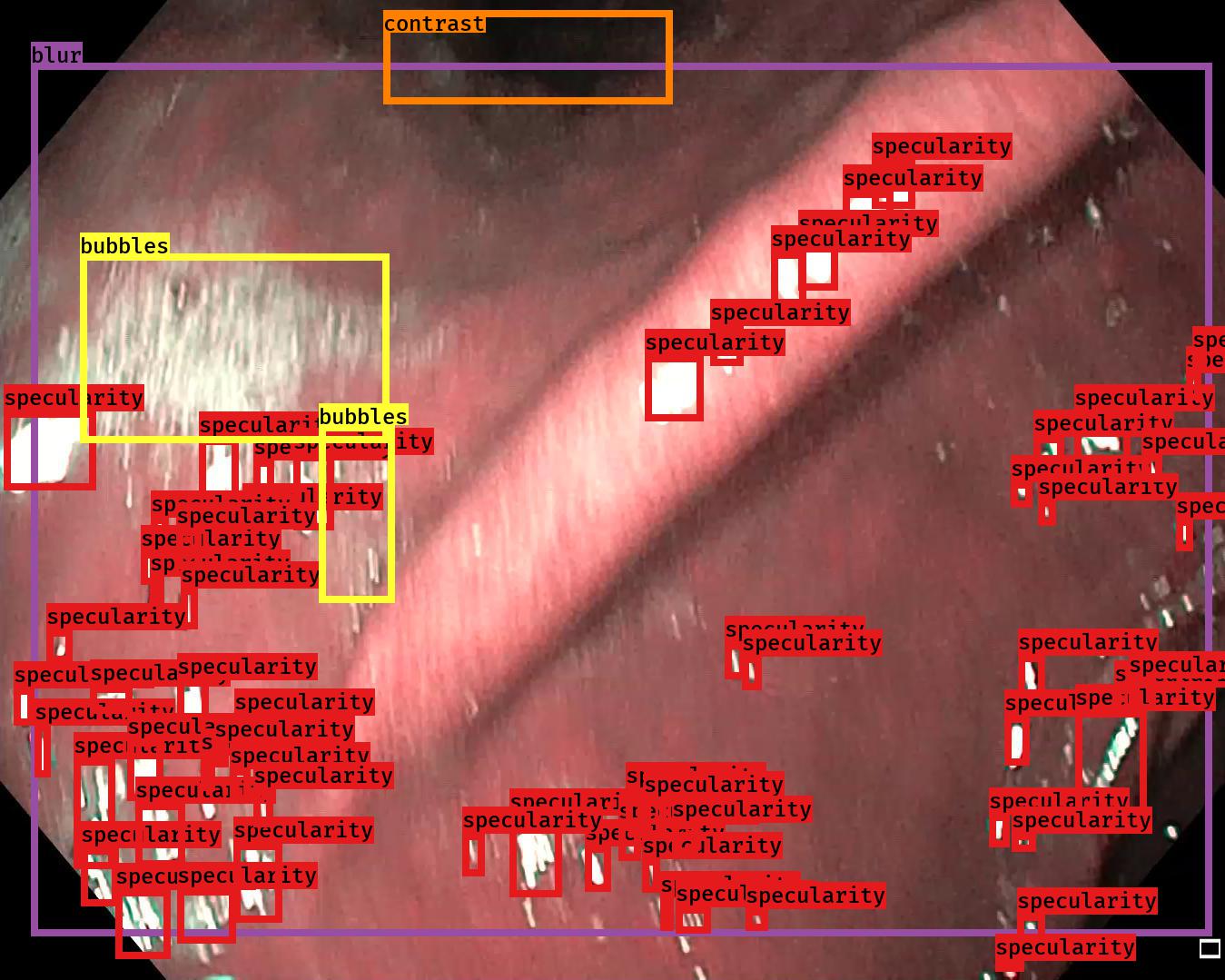}
\end{minipage}
\begin{minipage}[b]{1\linewidth}
\includegraphics[width=\linewidth]{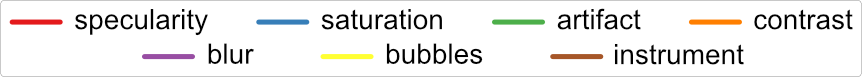}
\end{minipage}
\caption{Example annotated training detection boxes illustrating the 7 different artifact classes in the EAD2019 challenge dataset.}
\label{fig:example-detection-boxes}
\end{figure}

\begin{figure}[t!]
\centering
\begin{minipage}[b]{0.285\linewidth}
\includegraphics[width=\linewidth]{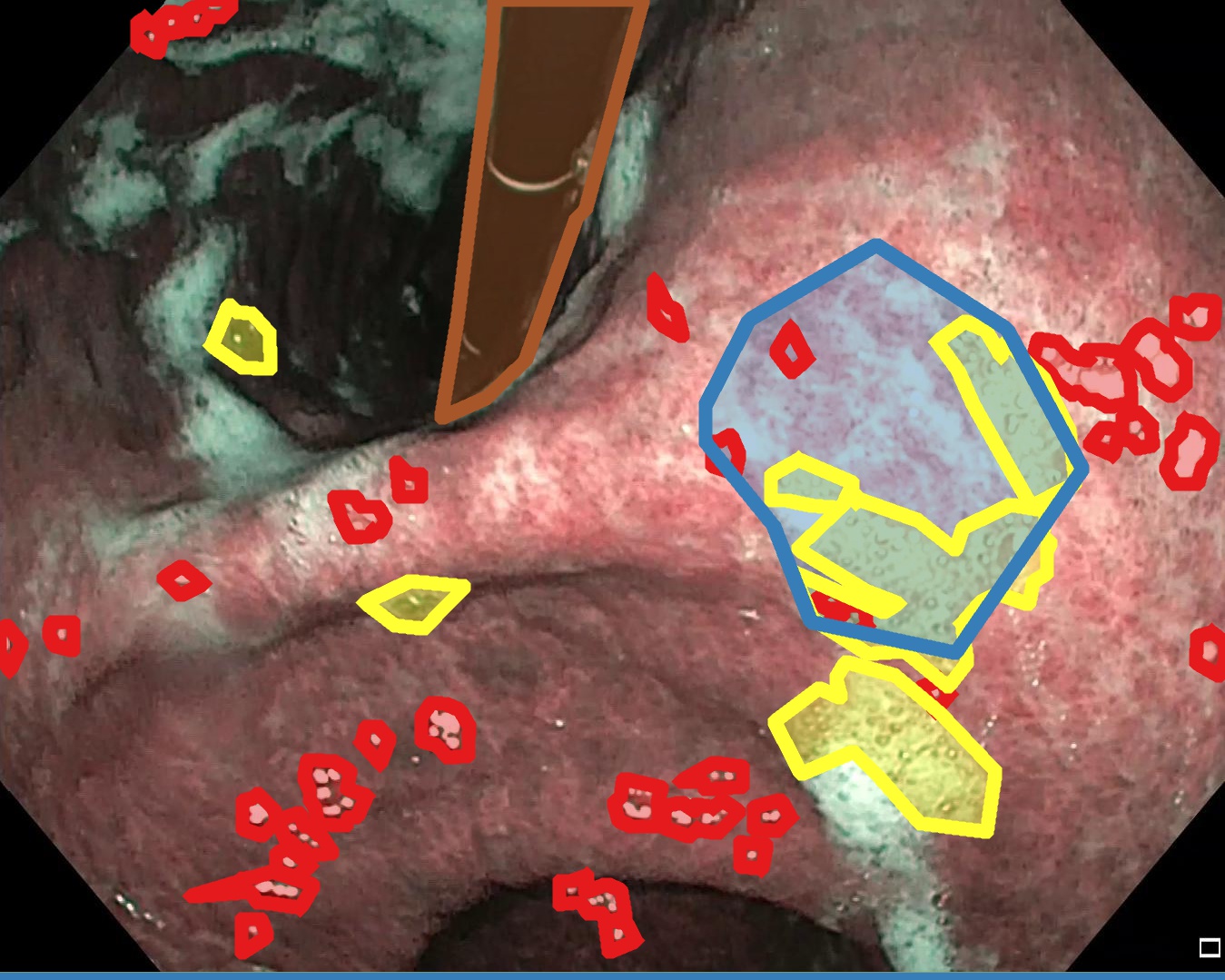}
\end{minipage}
\begin{minipage}[b]{0.285\linewidth}
\includegraphics[width=\linewidth]{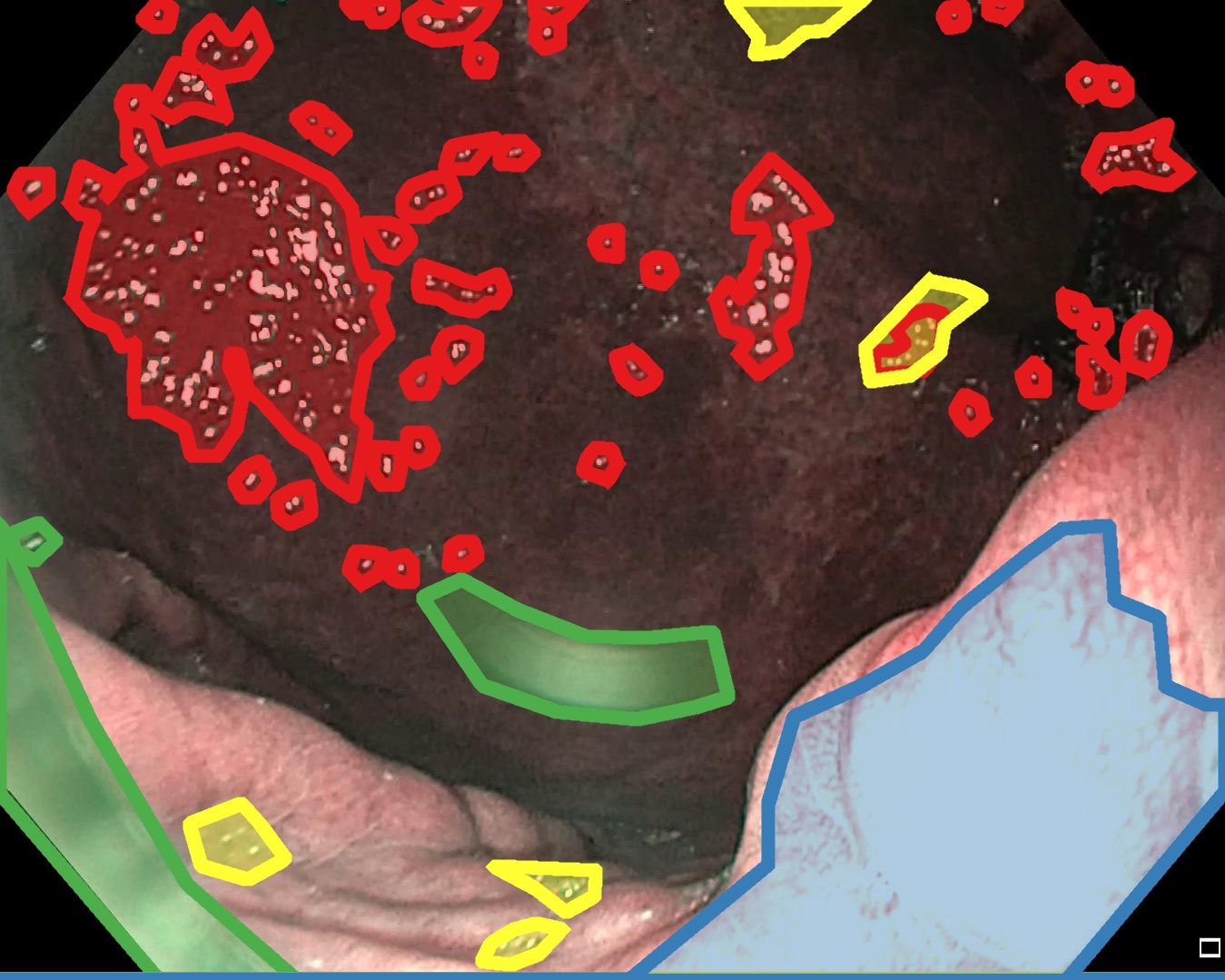}
\end{minipage}
\begin{minipage}[b]{0.24\linewidth}
\includegraphics[width=\linewidth]{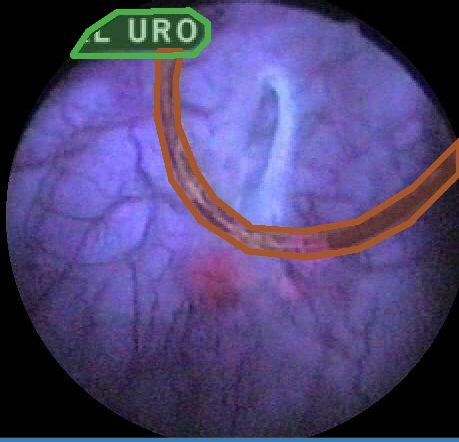}
\end{minipage}
\begin{minipage}[b]{0.32\linewidth}
\includegraphics[width=\linewidth]{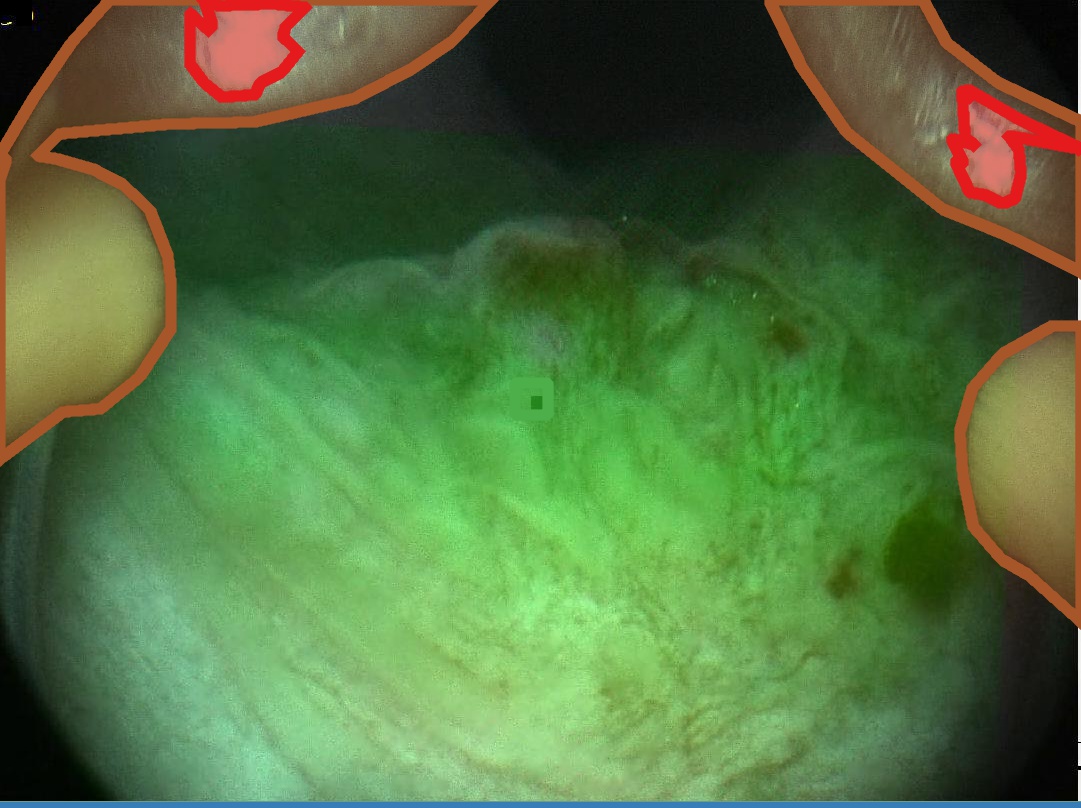}
\end{minipage}
\begin{minipage}[b]{0.24\linewidth}
\includegraphics[width=\linewidth]{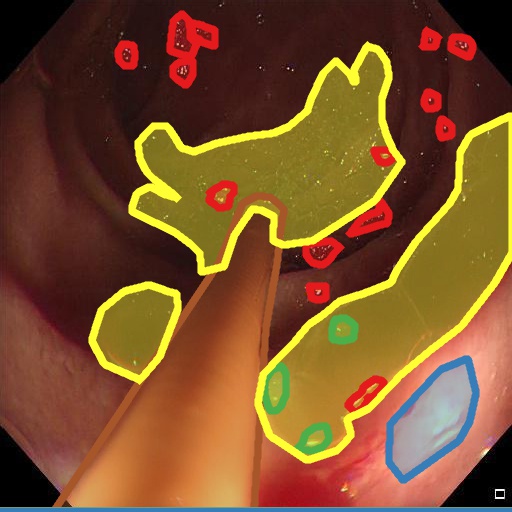}
\end{minipage}
\begin{minipage}[b]{1\linewidth}
\includegraphics[width=\linewidth]{legend-bar_ead2019.png}
\end{minipage}
\caption{Example of annotated segmentation masks. For segmentation only 5 of the 7 classes were annotated, specularity, bubbles, saturation, artifact and instrument}
\label{fig:example-segmentation-masks}
\end{figure}
\paragraph{Annotation software}
We have used an open-source VIA annotation tool for semantic segmenation~\cite{dutta2019vgg}. For bounding box annotation we have used a python, Qt and Opencv based in-house tool. 
\paragraph{Annotation Protocols}
\begin{itemize}
    \item For the same region, multiple boxes were annotated if the region belonged to more than 1 class 
    \item The minimal box sizes were used to describe the artifact region, e.g. if there are lots of specular reflections present in an image then instead of one large box we use multiple small boxes to capture the natural size of the artifact
    \item Each artifact type was determined to be distinctive and general across endoscopy datasets 
\end{itemize}
\paragraph{Annotator Variation}
\begin{itemize}
    \item Variation in bounding box annotations are considered by weighting the final score in the multi-class artifact detection challenge (0.6*mAP + 0.4*IoU) as IoU (intersection over union) is likely to vary more compared to mAP (mean average precision) across individual annotators
    \item Variation in the semantic class labels of masks for semantic segmentation was found not significant. Further we do not consider contrast and blur classes which are poorly defined.  
\end{itemize}
Examples for bounding box annotations for detection are shown in Fig.~\ref{fig:example-detection-boxes}. It can be observed that while multiple boxes are annotated for several small specular areas; contrast, blur and instrument have relatively larger areas. Due to the overlap between two or more classes, the annotation by experts varied. This was minimized by following the detailed annotation protocol above. For semantic segmentation, a larger area mask was preferentially used to delineate locally very cluttered small specularity artifacts (see Fig.~\ref{fig:example-segmentation-masks}).
\subsection{Training and testing data}
\begin{figure}[t!]
	\centering
	\begin{minipage}[b]{0.5\linewidth}
            \includegraphics[trim=.25cm 0.20cm 0.5cm 0.5cm, clip=true,scale=0.45]{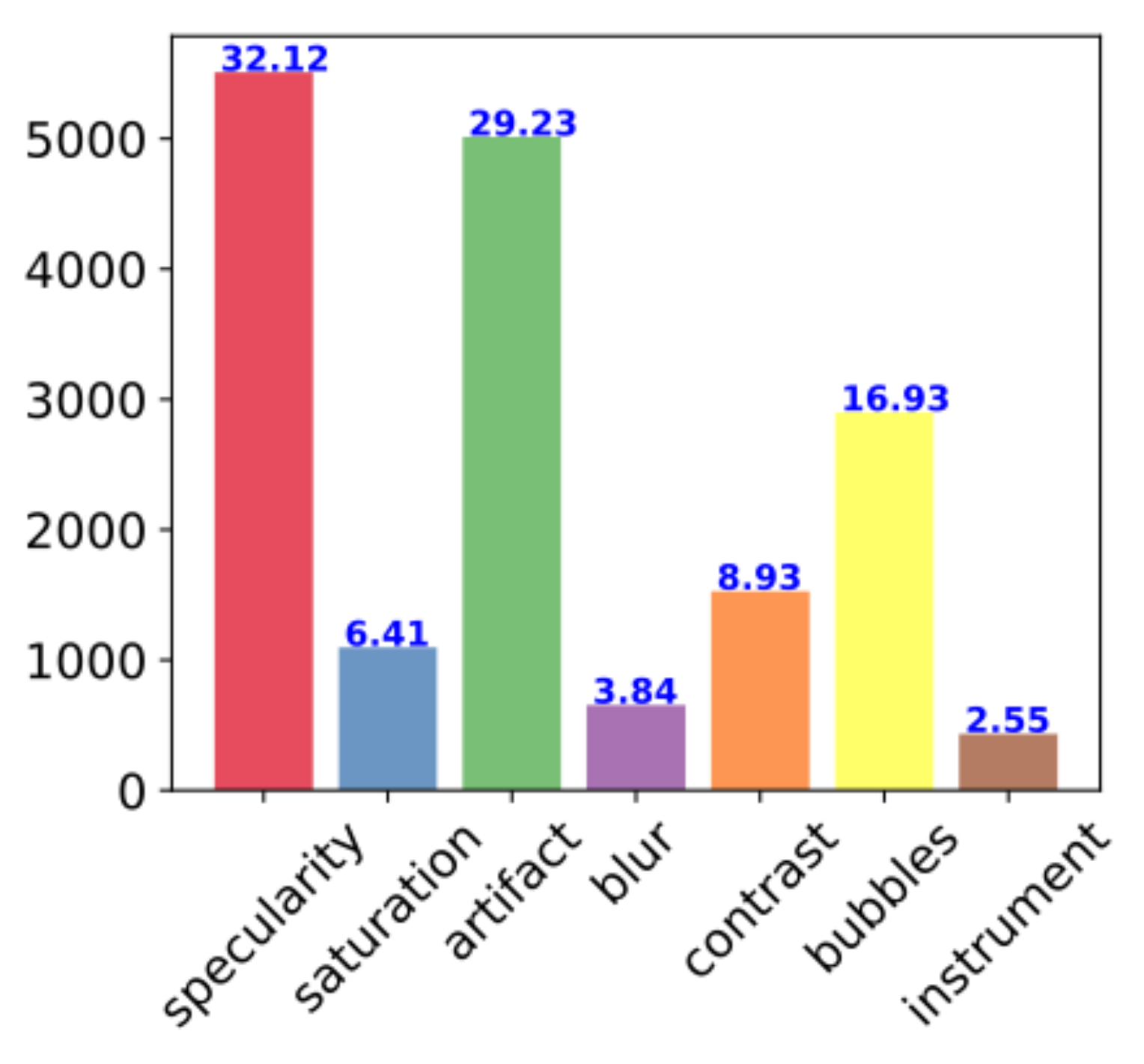}
            \centerline{\footnotesize{a. Training data for detection}}
    \end{minipage}
	\begin{minipage}[b]{0.48\linewidth}
        \includegraphics[trim=0.25cm 0.20cm 0.5cm 0.5cm, clip=true,scale=0.45]{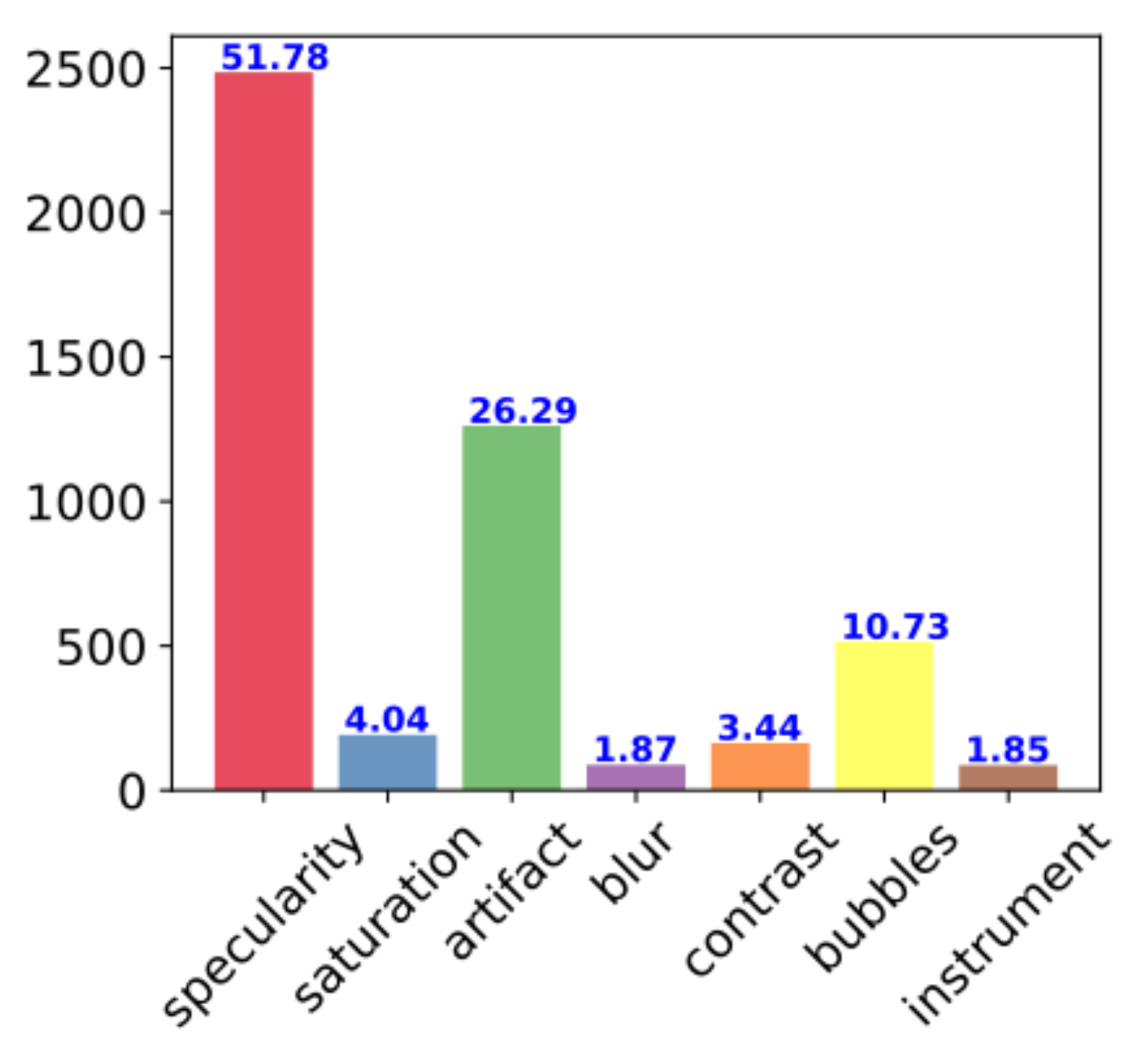}
        \centerline{\footnotesize{b. Test data for detection}}
    \end{minipage}
    
	\begin{minipage}[b]{0.5\linewidth}
        \includegraphics[trim=0.25cm 0.20cm 0.5cm 0.5cm, clip=true,scale=0.45]{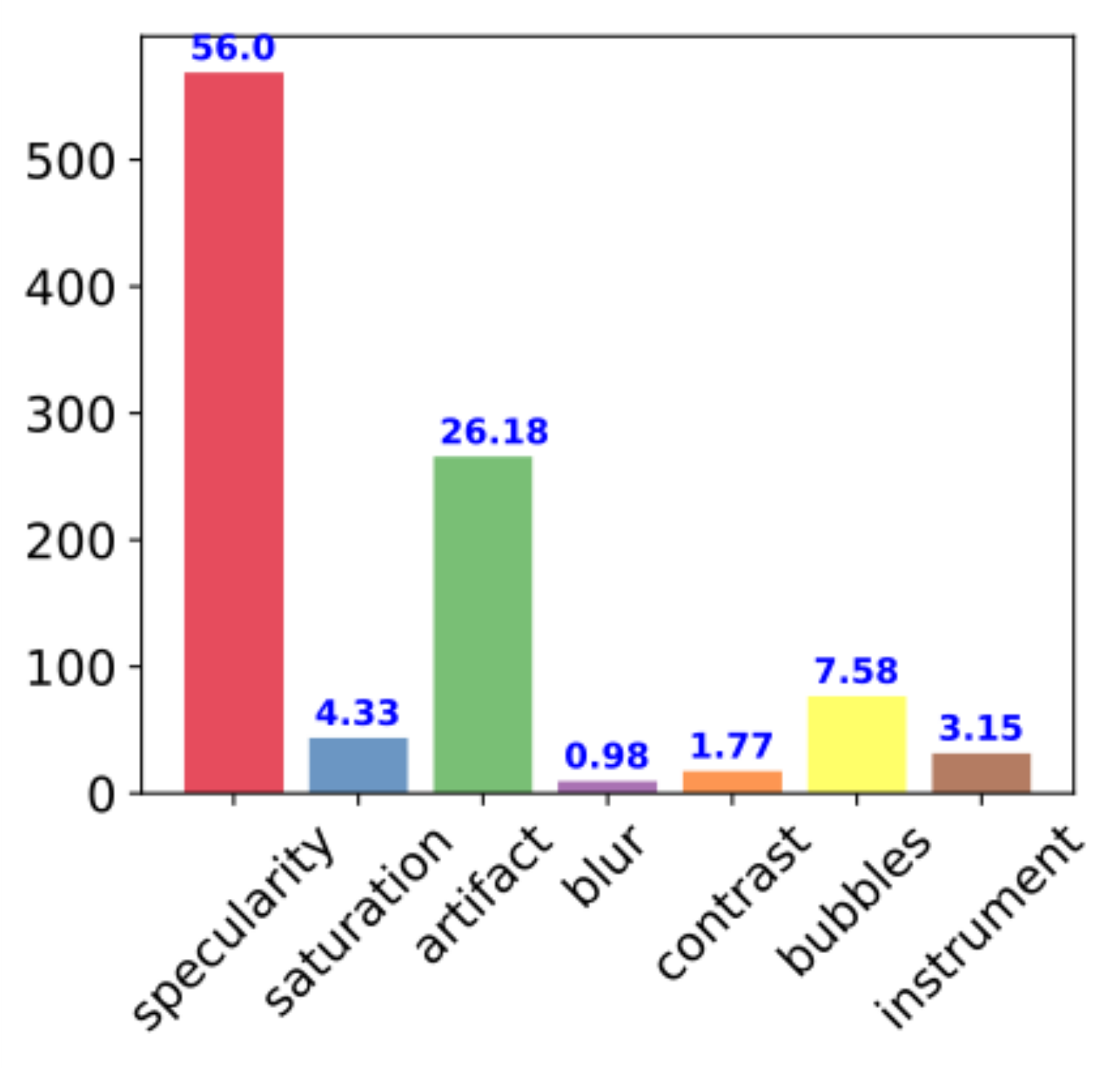}
            \centerline{\footnotesize{c. Test data for generalization}}
    \end{minipage}
    \caption{Artifact class distribution in train-test dataset for detection and generalization tasks. Top text in blue indicates the percentage of each class present in the individual dataset.}
    \label{fig:ead2019_datasummary_histograms}
\end{figure}
\subsubsection*{Detection}
The training dataset for detection consists in total 2147 annotated frames over all 7 artifact classes. All algorithms were evaluated online\footnote{https://ead2019.grand-challenge.org/evaluation/results/} using the evaluation metrics discussed in Section~\ref{sec:evalCriteria} on a test set of 195 frames ($\sim$10\% of training data). During the annotation we found that most of frames were much more affected by specularity, imaging artifact and bubbles compared to other artifact classes. We tried to keep the ratio of class types similar between the training and test datasets as best we could. Artifact class distribution for detection and generalization datasets are provided in Fig.~\ref{fig:ead2019_datasummary_histograms}.

\subsubsection*{Semantic Segmentation}
For semantic segmentation, we released 475 annotated frames for 5 different classes that include specularity, saturation, artifact, bubbles and instrument. For test data, 122 frames were annotated for online evaluation of participants algorithm. All data are available online at \citep{EAD2019Dataset}.

\subsubsection*{Generalization}
The training dataset for generalization is the same as that for detection however the test data for generalization uses a previously withheld dataset provided by a sixth institution not present in any other training or test data released for the detection and segmentation tasks. The generalization test data consisted of 52 images and the task was to detect all 7 artifact classes as with the detection task. 

\begin{figure}[t!]
	\centering
	\begin{minipage}[b]{0.5\linewidth}
            \includegraphics[scale=0.45]{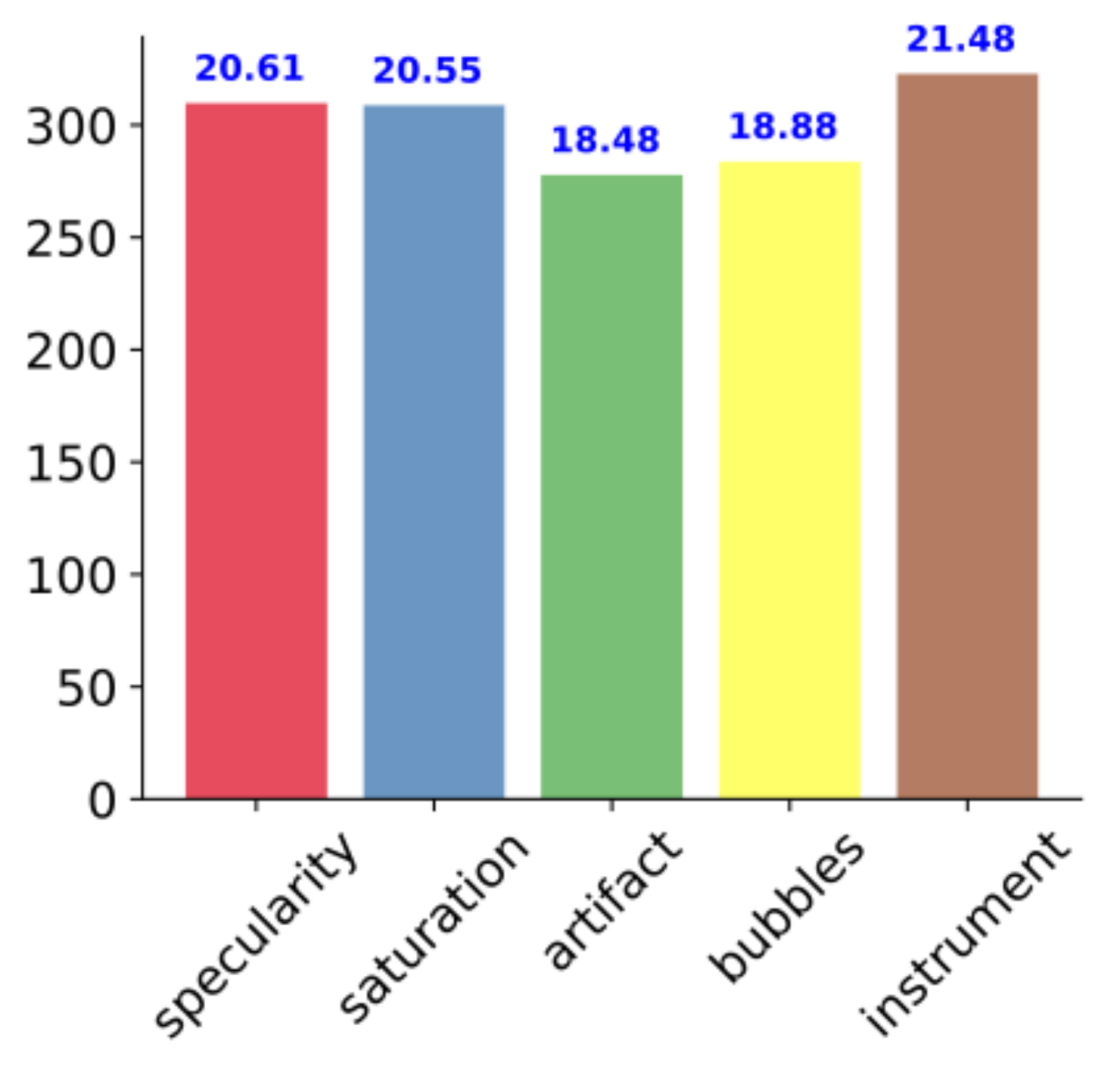}
    \end{minipage}
	\begin{minipage}[b]{0.48\linewidth}
        \includegraphics[scale=0.45]{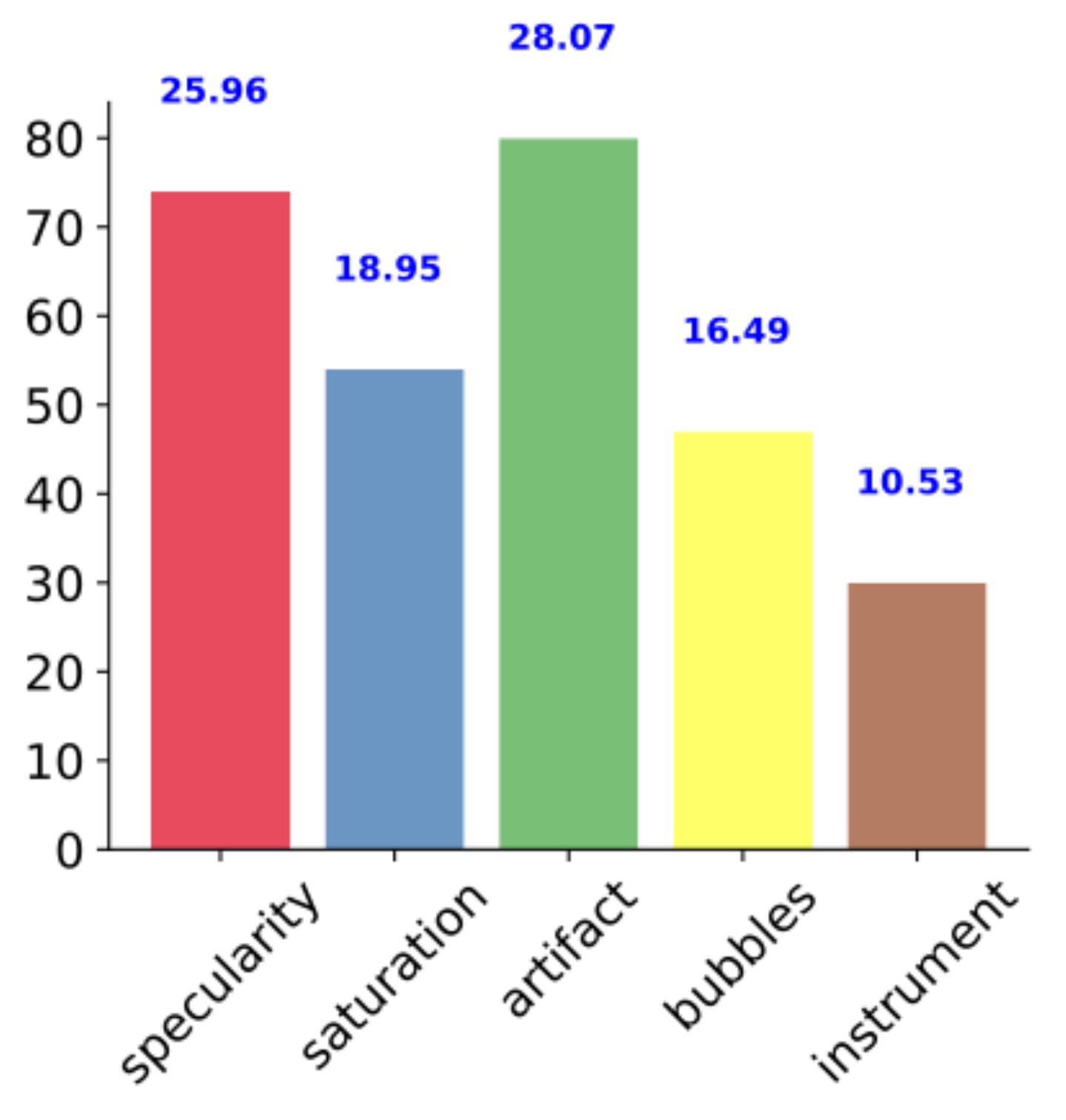}
    \end{minipage}
    \caption{Train and test data distribution per class for semantic segmentation. Left: Train data and right: Test data. Top text in blue indicates the percentage of each class present in the individual dataset.}
    \label{fig:ead2019_datasummary_histogram_semantic}
\end{figure}
%
\section{Evaluation criteria}{\label{sec:evalCriteria}}
The challenge problems fall into three distinct categories. For each there exists already well-defined evaluation metrics used by the wider imaging community which we use for evaluation here. 
\subsection{Detection score}
\begin{figure}[t!]
	\centering
        \includegraphics[width=1\linewidth]{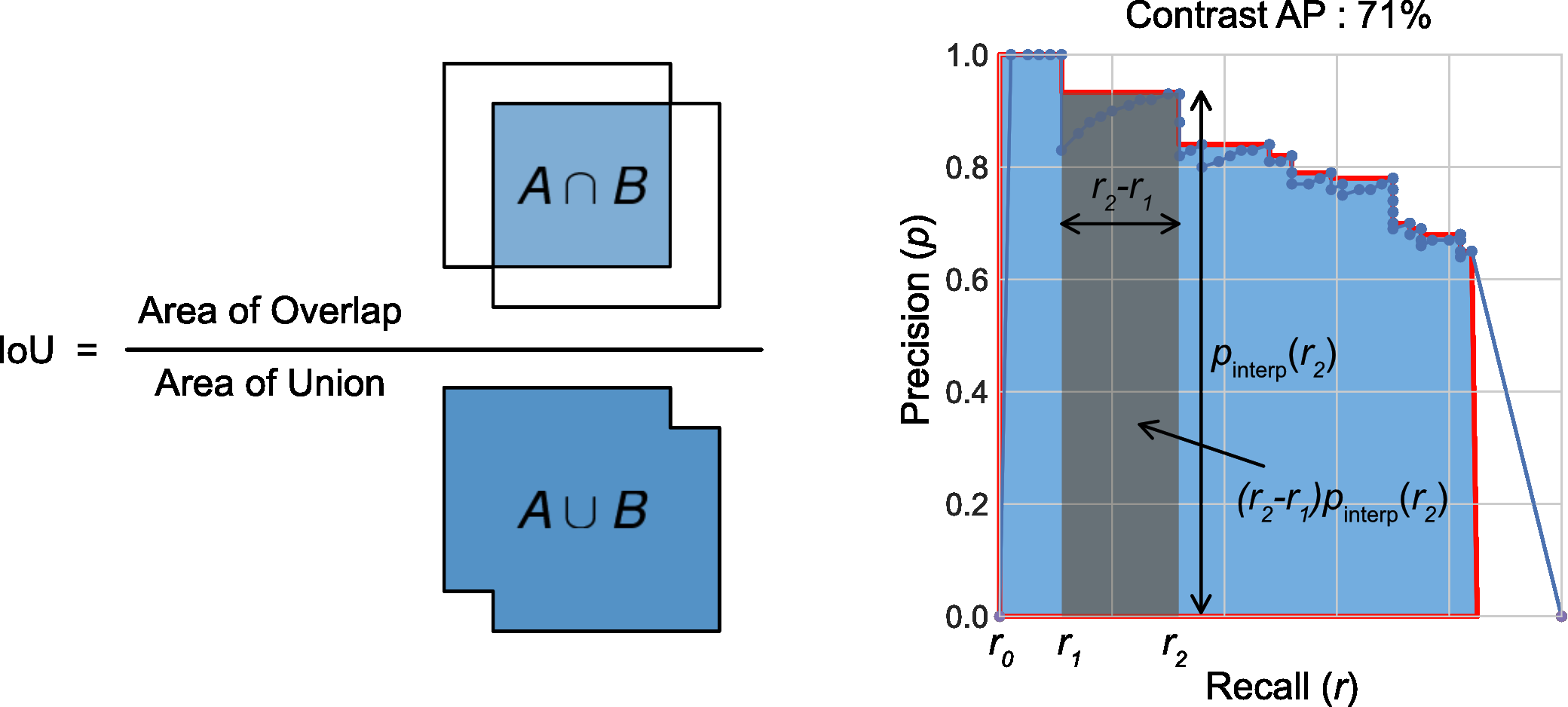}
    \caption{Schematic illustration of detection evaluation metrics. Left: intersection over union (IoU) defined for two rectangles A and B. Right: mean average precision (mAP) is the mean AP over all 7 artifact classes where AP (blue area) is computed as the sum of the individual rectangular areas (grey area) under the precision-recall curve sampled at all unique recall values ($r_n$).} 
    \label{fig:detection_metrics_illustration}
\end{figure}

Metrics used for multi class artifact detection: 
\begin{itemize}
    \item IoU - intersection over union. This metric measures the overlap between two bounding boxes $A$ and $B$ as the ratio between the overlapped area $A\cap B$ over the total area $A\cup B$ occupied by the two boxes (see Fig.\ref{fig:detection_metrics_illustration}):
    \begin{equation}
        \mathrm{IoU} = \frac{A\cap B}{A\cup B}
    \end{equation}
    where $\cap$, $\cup$ denote the intersection and union respectively. In terms of numbers of true positives (TP), false positives (FP) and false negatives (FN), IoU (aka Jaccard J) can be defined as:
    \begin{equation}
        IoU/J = \frac{TP}{TP+FP+FN}
    \end{equation}
    
    \item mAP - mean average precision of detected artifacts with precision (p) defined as $p={\frac {TP}{TP+FP}}$ and recall (r) as $r={\frac {TP}{TP+FN}}$. {This} metric measures the ability of an object detector to accurately retrieve all instances of the ground truth bounding boxes. The higher the mAP the better the performance. Average precision (AP) is computed as the Area Under Curve (AUC) of the precision-recall curve of detection sampled at all unique recall values $(r_1, r_2,...)$ whenever the maximum precision value drops:
    \begin{equation}
        \mathrm{AP} = \sum_n{\left\{\left(r_{n+1}-r_{n}\right)p_{\mathrm{interp}}(r_{n+1})\right\}}, 
    \end{equation}

    with $p_{\mathrm{interp}}(r_{n+1}) =\underset{\tilde{r}\ge r_{n+1}}{\max}p(\tilde{r})$. Here, $p(r_n)$ denotes the precision value at a given recall value. This definition ensures monotonically decreasing precision. The mAP is the mean of AP over all artifact classes $i$ for $N=7$ classes given as
    \begin{equation}
        \mathrm{mAP} = \frac{1}{N}\sum_i{\mathrm{AP}_i}
    \end{equation}
    This definition was popularised in the PASCAL VOC challenge~\cite{pascal-voc-2012}. The calculation is illustrated in Fig. \ref{fig:detection_metrics_illustration}. An IoU$>=0.25$ was used to call a "match" between a predicted and ground-truth detection.
\end{itemize}
Participants were finally ranked on a final mean score $(\mathrm{score_d})$, a weighted score of mAP and IoU represented as:
\begin{equation}
\mathrm{score_d} = 0.6 \times \mathrm{mAP_d} + 0.4 \times \mathrm{IoU_d}
\end{equation}
\subsection{Semantic segmentation score}
Metrics widely used for multi-class semantic segmentation of artifacts have been used for scoring semantic segmentation. It comprises of: 
\begin{itemize}
    \item Dice similarity coefficient (DSC) or F1-score 
    \item Jaccard Index (J) or IoU
    \item F2-error
\end{itemize}
The general forumula to compute $\mathrm{F_\beta}$-score is:
\begin{equation}{\label{eq:fscore}}
    F_\beta = (1 + \beta^2) \cdot \frac{{p} \cdot {r}}{(\beta^2 \cdot {p}) + {r}}
\end{equation}
With $\beta = 1$ and $\beta = 2$ in Eq.~(\ref{eq:fscore}) one can compute F1-score (DSC) and F2-error respectively. Participants were ranked on a final weighted score $(\mathrm{score_s})$ of the above metrics defined as:
\begin{equation}
\mathrm{score_s} = 0.75 \times \left[0.5 \times\mathrm{(DSC + J)}\right] + 0.25 \times \mathrm{F2}-\mathrm{error}
\end{equation}
\subsection{Generalization score}
For multi-class artifact detection, task-3 generalization detection mAP was estimated on a sixth institution dataset not included in the training or test data of the detection and segmentation tasks. Generalization was evaluated based on the deviation between the mAP of the detection and generalization test datasets for strictly the same model parameters. Participants were ranked based on a score gap of generalization defined as:
\begin{equation}
    \mathrm{dev_g} = \mathrm{mAP_d} - \mathrm{mAP_g}
\end{equation}
It is worth noting that the highest $\mathrm{mAP_g}$ with $\mathrm{dev_g} \rightarrow 0$ is required for winning the competition.

\section*{Acknowledgement}
\noindent{We} would like to acknowledge James Meakin and his team for providing us an online framework to host our challenge at grand-challenges.org. We would also like to thank the Medical Image Analysis Network (MedIAN) and Cancer Research UK (CRUK) for co-sponsoring our workshop event at IEEE ISBI 2019, Venice, Italy. We are grateful to our colleagues especially Mariia Dmitrieva, Korsuk Sirinukunwattana, Soumya Gupta, Ka Ho Tam and Joel Lefebvre who helped during our annotation protocol study.  
\bibliographystyle{elsarticle-num}
\bibliography{sample.bib}
\end{document}